# A Hybrid Feature Selection Method to Improve Performance of a Group of Classification Algorithms

Mehdi Naseriparsa
Islamic Azad University
Tehran North Branch
Dept. of Computer Engineering
Tehran, Iran

Amir-Masoud Bidgoli
Islamic Azad University
Tehran North Branch
MIEEE Manchester University
Tehran, Iran

Touraj Varaee
Islamic Azad University
Tehran North Branch
Dept. of Computer Engineering
Tehran, Iran

## ABSTRACT
In this paper a hybrid feature selection method is proposed which takes advantages of wrapper subset evaluation with a lower cost and improves the performance of a group of classifiers. The method uses combination of sample domain filtering and resampling to refine the sample domain and two feature subset evaluation methods to select reliable features. This method utilizes both feature space and sample domain in two phases. The first phase filters and resamples the sample domain and the second phase adopts a hybrid procedure by information gain, wrapper subset evaluation and genetic search to find the optimal feature space. Experiments carried out on different types of datasets from UCI Repository of Machine Learning databases and the results show a rise in the average performance of five classifiers (Naïve Bayes, Logistic, Multilayer Perceptron, Best First Decision Tree and JRIP) simultaneously and the classification error for these classifiers decreases considerably. The experiments also show that this method outperforms other feature selection methods with a lower cost.

## 1. INTRODUCTION
The term data mining refers loosely to the process of semi automatically analyzing large databases to find useful patterns [1]. Like knowledge discovery in artificial intelligence (also called machine learning) or statistical analysis, data mining attempts to discover rules and patterns from data [1]. High dimensional datasets usually lead to deteriorate the accuracy and performance of the system by curse of dimensionality. Datasets with high dimensional features have more complexity and spend longer computational time for classification [2]. Feature selection is a solution to high dimensional data. Feature selection is an important topic in data mining, specifically for high dimensional datasets. Feature selection is a process commonly used in machine learning, wherein subsets of the feature available from the data are selected for application of a learning algorithm. The best subset contains the least number of dimensions that most contribute to accuracy [3]. Feature selection aims to improve machine learning performance [4].

Janecek[5] showed the relationship between feature selection and data classification and the impact of applying PCA on the classification process. Assareh[6] proposed a hybrid random subspace fusion model that utilizes both the feature space and sample domain to improve the diversity of the classifier ensemble. Hayward[7] showed that data preprocessing and choosing suitable features will develop the performance of classification algorithms. Dhiraj[8] used clustering and K-means algorithm to show the efficiency of this method on huge amount of data. Xiang[9] proposed a hybrid feature selection algorithm that takes the benefit of symmetrical uncertainty and genetic algorithms. Zhou[10] presented a new approach for classification of multi class data. The algorithm performed well on two kind of cancer datasets. Fayyad[11] tried to adopt a method to seek effective features of dataset by applying a fitness function to the attributes.

Most of the feature selection methods work on feature space and they do not test the effect of filtering and resampling instances on the feature selection process. Furthermore, feature selection methods usually focus on one specific classification algorithm to test their performance. Hence, only one part of the sample space patterns are covered [6]. In this paper, we try to test our proposed feature selection method on a group of classification algorithms. In fact, our proposed method takes the advantages of combining sample domain filtering, resampling and feature subset evaluation methods to improve the performance of a group of classification algorithms simultaneously. Since feature selection methods are designed for high dimensional datasets, their performance on small and middle sized datasets is not acceptable. Hence, we also try to propose an adaptive feature selection method that is applicable for most of datasets with different sizes.

In section 2, 3, 4, 5, 6, 7 and 8 we focus on the definition of feature selection, SMOTE, wrapper approach, Naïve Bayes classifier, entropy measure, information gain and genetic algorithm which are used in our proposed method. In section 9, we describe our hybrid method and explain the two phases involved in the feature selection process. In section 10, the performance of the proposed method is tested on various datasets. Conclusions are given in section11.

## 2. FEATURE SELECTION
Feature selection can be defined as a process that chooses a minimum subset of M features from the original set of N features, so that the feature space is optimally reduced according to a certain evaluation criterion [12]. As the dimensionality of a domain expands, the number of feature N increases. Finding the best feature subset is usually intractable [13] and many problems related to feature selection have been shown to be NP-hard [14]. Researchers have studied various aspects of feature selection. Feature selection algorithms may be divided into filters [15], wrappers [13] and embedded approaches [16]. Filters method evaluate quality of selected features, independently from the classification algorithm, while wrapper methods require application of a classifier to evaluate this quality. Embedded methods perform feature selection during learning of optimal parameters.

## 3. SMOTE: SYNTHETIC MINORITY OVER-SAMPLING TECHNIQUE

Often real world datasets are predominantly composed of normal examples with only a small percentage of abnormal or interesting examples. It is also the case that the cost of misclassifying an abnormal example as a normal example is often much higher than the cost of the reverse error. Under sampling of the majority (normal) class has been proposed as a good means of increasing the sensitivity of a classifier to the minority class. By combination of over-sampling the minority (abnormal) class and under-sampling the majority (normal) class, the classifiers can achieve better performance than only under-sampling the majority class. SMOTE adopts an over-sampling approach in which the minority class is over-sampled by creating synthetic examples rather than by over-sampling with replacement. The synthetic examples are generated in a less application specific manner, by operating in feature space rather than sample domain. The minority class is over-sampled by taking each minority class sample and introducing synthetic examples along the line segments joining any of the k minority class nearest neighbors. Depending upon the amount of over-sampling required, neighbors from the k nearest neighbors are randomly chosen [17].

## 4. WRAPPER APPROACH

In the wrapper approach, the feature subset selection is done using the induction algorithm as a black box. The feature subset selection algorithm conducts a search for a good subset using the induction algorithm itself as part of the evaluation function. The accuracy of the induced classifiers is estimated using accuracy estimation techniques [18]. Wrappers are based on hypothesis. They assign some values to weight vectors, and compare the performance of a learning algorithm with different weight vector. In wrapper method, the weights of features are determined by how well the specific feature settings perform in classification learning. The algorithm iteratively adjust feature weights based on its performance.

The induction algorithm in Wrapper method could be Naïve Bayes classifier [18]. In this algorithm, selective Bayesian which uses a forward and backward greedy search method is applied to find a feature subset from the whole space of entire features. It uses the accuracy of Naïve Bayes classifier on the training data to evaluate feature subsets, and considers adding each unselected feature which can improve the accuracy on each iteration. The method shows a significant improvement over Naïve Bayes. However, a major disadvantage associated with the wrapper mechanism is the computational cost involved.

## 5. NAIVE BAYES

The Naive Bayes algorithm is based on conditional probabilities. It uses Bayes' Theorem, a formula that calculates a probability by counting the frequency of values and combinations of values in the historical data. Bayesian classifiers find the distribution of attribute values for each class in the training data [1]. When given a new instance d, they use the distribution information to estimate, for each class cj, the probability that instance d belongs to class cj, denoted by p(cj | d). The class with maximum probability becomes the predicted class for instance d. to find the probability p(cj | d) of instance d being in class cj, Bayesian classifiers use Bayes theorem as shown in equation(1).

$$P(C_j \mid d) = \frac{P(d \mid c_j) P(c_j)}{P(d)} \quad (1)$$

Where P(d | cj) is the probability of generating instance d given class cj, P(cj) is the probability of occurrence of class cj, and P(d) is the probability of instance d occurring. The Naive Bayes classifier is designed for use when features are independent of one another within each class, but it appears to work well in practice even when that independence assumption is not valid. Thereby it estimates P(d|cj ) as shown in equation(2).

$$P(d \mid c_j) = P(d_1 \mid c_j) \times P(d_2 \mid c_j) \times \ldots \times P(d_n \mid c_j) \quad (2)$$

Naïve Bayes classifier learns from training data from the conditional probability of each attribute given the class label. Using Bayes rule to compute the probability of the classes given the particular instance of the attributes, prediction of the class is done by identifying the class with the highest posterior probability. Computation is made possible by making the assumption that all attributes are conditionally independent given the value of the class. Naïve Bayes as a standard classification method in machine learning stems partly because it is easy to program, its intuitive, it is fast to train and can easily deal with missing attributes. Research shows Naïve Bayes still performs well in spite of strong dependencies among attributes.

## 6. ENTROPY MEASURE

Entropy is a common measure in the information theory, which characterizes the purity of an arbitrary collection of examples [19]. The entropy measure is considered as a measure of system's unpredictability. The entropy of Y is shown in equation 3:

$$H(Y) = - \sum_{y \in Y} p(y) \log_2(p(y)) \quad (3)$$

Where p(y) is the marginal probability density function for the random variable Y. If the observed values of Y in the training dataset S are partitioned according to the values of a second feature X, and the entropy of Y with respect to the partitions induced by X is less than the entropy of Y prior to partitioning, then there is a relationship between features Y and X. Then the entropy of Y after observing X is shown in equation 4:

$$H(Y \mid X) = - \sum_{x \in X} p(x) \sum_{y \in Y} p(y \mid x) \log_2(p(y \mid x)) \quad (4)$$

Where p(y|x) is the conditional probability of y given x.

## 7. INFORMATION GAIN
The information gain of a given attribute X with respect to the class attribute Y is the reduction in uncertainty about the value of Y when we know the value of X. The uncertainty about the value of Y is measured by its entropy, H(Y). The uncertainty about the value of Y when we know the value of X is given by the conditional entropy of Y given X, H(Y|X) as shown in 5:

$$IG = H(Y) - H(Y|X) = H(X) - H(X|Y) \quad (5)$$

IG is a symmetrical measure [12]. The information gained about Y after observing X is equal to the information gained about X after observing Y.

## 8. GENETIC ALGORITHMS
The genetic algorithm is a method for solving both constrained and unconstrained optimization problems that is based on natural selection, the process that drives biological evolution [20]. Genetic Algorithms are a family of computational models inspired by evolution. Theses algorithms encode a potential solution to a specific problem on a simple chromosome-like data structure and apply recombination operators to these structures so as to preserve critical information. Genetic Algorithms are often viewed as function optimizers, although the range of problems to which Genetic Algorithms have been applied is quite broad.

## 9. PROPOSED METHOD
### 9.1 Initial Phase
In the first phase, the sample domain analysis is performed and a secondary dataset is derived from the original dataset. In the first step, the SMOTE technique is applied on the original dataset to increase the samples of the minority class. This step contributes to make a more diverse and balanced dataset. Although resampling techniques like SMOTE help to produce a more reliable and diverse dataset, they may deteriorate the classification performance due to addition of unreliable samples. To avoid the deleterious effect of such samples in the second step, sample domain filtering method is applied on the resulting dataset to refine the dataset and omit the unreliable samples which are misclassified by the learning algorithm. The learning algorithm for filtering is Naïve Bayes. Naïve Bayes eliminates misclassified samples which are added to the dataset during the resampling process by a low computational cost. This step is crucial because some samples which are produced by SMOTE may mislead the classifiers and result to lower the performance of classification.

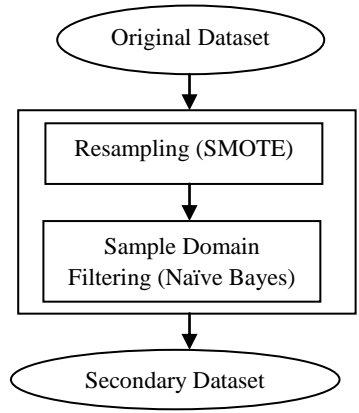

**Figure 1: Flow diagram of first step in the first phase**

Finally, the original dataset is merged with the secondary dataset. The resulting dataset keeps all the samples of the original dataset and also has some additional samples which contribute to improve accuracy and performance of a group of classification algorithms.

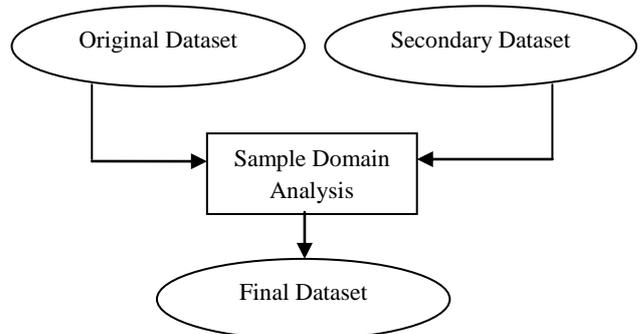

**Figure 2: Combination of original dataset with the secondary dataset**

### 9.2 Secondary Phase
In the second phase, the feature space is searched to reach the best subset that results in the best accuracy and performance for the group of classification algorithms. Actually, feature space analysis is carried out in two steps. In the first step, a feature space filtering method is adopted to reduce the feature space and prepare the conditions for the next step. Information gain is a filtering method which uses entropy metric to rank the features and is used for the first filtering step. At the end of this step the features with the ranks higher than the threshold are selected for the next round.

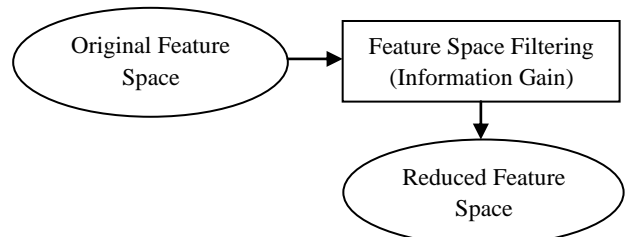

**Figure 3: Flow diagram of first step in the second phase**

In the second step, wrapper feature selection with genetic search is carried out on the remaining feature subset. Naïve Bayes is chosen as the learning algorithm for wrapper feature selection. The initial population for genetic search is set by the order of features which has been defined by Information gain in the previous step. The features are chosen at the end of this phase are considered as the reliable features.

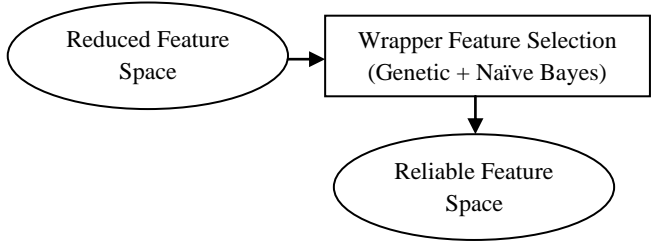

**Figure 4: Flow diagram of second step in the second phase**

Different experiments are carried out on various datasets with a group of classification algorithms and the results exhibit that the average performance of the classification algorithms improved in comparison with other feature selection methods.

Wrapper feature selection is a costly method due to its comprehensive search on the feature space. Hence, seeking reliable features using wrapper is impossible for the datasets with huge feature space. In order to take advantages of this highly accurate method and also reducing its computational cost, we used a hybrid strategy in the second phase of the proposed method. In fact, the first step of the feature space reduction uses information gain filtering to eliminate the unrelated and redundant features before the application of the costly wrapper method. This step reduces the feature space considerably and consequently the wrapper method is carried out on a smaller space which is more efficient.

## 10. EMPIRICAL STUDY
### 10.1 Experimental Setup and Conditions

To evaluate our feature selection method, we choose 5 datasets from UCI Repository of Machine Learning databases [21] and apply 5 important classification algorithms before and after implementation of our feature selection method. A summary of datasets are presented in table 1.

**Table 1. Characteristics of UCI datasets used in experiments**

| Dataset Name | Samples | Features | Classes |
|---|---|---|---|
| Lung-Cancer | 32 | 56 | 3 |
| WDBC | 569 | 30 | 2 |
| Hepatitis | 155 | 19 | 2 |
| Dermatology | 366 | 34 | 6 |
| Wine | 178 | 13 | 3 |

GA parameters are set as follows: Crossover Probability is the probability that two population members will exchange genetic material and is set to 0.6. Max Generations parameter show the number of generations to evaluate and is set to 20. Mutation Probability is the probability of mutation occurring and is set to 0.033 and the last parameter is the number of individuals (attribute sets) in the population that is set to 20. The initial states for classification algorithms are the default state of WEKA software.

### 10.2 Performance Evaluation Parameters

The first parameter, is the average number of misclassified samples of the datasets on which the classifiers applied and we call it AMS. This parameter shows the efficiency of the feature selection method more realistically. AMS parameter formula is shown in equation 6 as:

$$AMS = \frac{\sum_{i=1}^{n} MS_i}{N} \quad (6)$$

In equation 6, $MS_i$ is the number of misclassified samples for a specific classification algorithm and N is the number of classification algorithms in the experiment. In table 2, the name and index of the classifiers are shown.

**Table 2. Name and index of classification algorithms**

| i | Classification Algorithm |
|---|---|
| 1 | Naïve Bayes |
| 2 | Logistic Regression |
| 3 | Multilayer Perceptron |
| 4 | BF Tree |
| 5 | JRIP |

The second parameter is the overall average misclassified samples and we call it OAMS. This parameter calculates the performance and accuracy of the feature selection method on a wider domain. Actually, the average misclassified samples of applying the classification algorithms on different datasets is estimated by this parameter. OAMS parameter formula is shown in equation 7 as:

$$OAMS = \frac{\sum_{i=1}^{n} AMS_i}{N} \quad (7)$$

In equation 7, $AMS_i$ is the average number of misclassified samples for a specific dataset on which a group of classification algorithms applied and N is the number of datasets used in the experiment.

The third parameter, is the Average relative absolute error [22] of the classification and we call it ARAE. This parameter shows how a feature selection method could affect the classifiers not to predict wrongly or at least their predictions are closer to the correct values. ARAE parameter formula is shown in equation 8 as:

$$ARAE = \frac{\sum_{i=1}^{n} RAE_i}{N} \quad (8)$$

In equation 8, RAE$_i$ is the relative absolute error for a specific classification algorithm and N is the number of classification algorithms in the experiment.

The next parameter is the overall average relative absolute error and we call it OARAE. This parameter calculates the relative absolute error of the feature selection method on a wider domain. OARAE formula is shown in equation 9 as :

$$OARAE = \frac{\sum_{i=1}^{n} ARAE_i}{N} \quad (9)$$

In equation 9, ARAE$_i$ is the average relative absolute error for a specific dataset on which a group of classification algorithms applied and N is the number of datasets used in the experiment.

Next parameters are about correctly classified rates [23]. True positive rate is the rate of correctly classified samples that belong to a specific class.

$$ATPRate = \frac{\sum_{i=1}^{n} TPRate_i}{N} \quad (10)$$

In equation 10, TPRate$_i$ is the true positive rate for a specific classification algorithm and N is the number of classification algorithms in the experiment.

The last parameter is the overall average true positive rate and we call it OATPRate. This parameter is shown in equation 11 as:

$$OATPRate = \frac{\sum_{i=1}^{n} ATPRate_i}{N} \quad (11)$$

In equation 11, ATPRate$_i$ is the average true positive rate for a specific dataset on which a group of classification algorithms applied and N is the number of datasets used in the experiment.

## 10.3 Experimental Results

To evaluate the performance of the proposed method, we choose 5 datasets with different sizes from both aspects of sample domain and feature space. The results of sample domain filtering and resampling in the first phase and feature space reduction for each dataset are presented in table3 to table7.

**Table 3. Results from running our proposed method on lung cancer dataset**

| Steps | Initial State | 1$^{st}$ Phase | 2$^{nd}$ Phase(1) | 2$^{nd}$ Phase(2) |
|---|---|---|---|---|
| Attributes | 56 | 56 | 28 | 12 |
| Samples | 32 | 93 | 93 | 93 |

**Table 4. Results from running our proposed method on WDBC dataset**

| Steps | Initial State | 1$^{st}$ Phase | 2$^{nd}$ Phase(1) | 2$^{nd}$ Phase(2) |
|---|---|---|---|---|
| Attributes | 32 | 32 | 19 | 11 |
| Samples | 596 | 1371 | 1371 | 1371 |

**Table 5. Results from running our proposed method on Hepatitis dataset**

| Steps | Initial State | 1$^{st}$ Phase | 2$^{nd}$ Phase(1) | 2$^{nd}$ Phase(2) |
|---|---|---|---|---|
| Attributes | 19 | 19 | 16 | 9 |
| Samples | 155 | 451 | 451 | 451 |

**Table 6. Results from running our proposed method on Dermatology dataset**

| Steps | Initial State | 1$^{st}$ Phase | 2$^{nd}$ Phase(1) | 2$^{nd}$ Phase(2) |
|---|---|---|---|---|
| Attributes | 34 | 34 | 28 | 20 |
| Samples | 366 | 1340 | 1340 | 1340 |

**Table 7. Results from running our proposed method on Wine dataset**

| Steps | Initial State | 1$^{st}$ Phase | 2$^{nd}$ Phase(1) | 2$^{nd}$ Phase(2) |
|---|---|---|---|---|
| Attributes | 13 | 13 | 12 | 10 |
| Samples | 178 | 415 | 415 | 415 |

According to the results of table3 to table7, we observe that this method leads to a good level of dimensionality reduction. In fact, the proposed method acts on two dimensions. At first, the method tries to refine the sample domain by resampling and filtering and then eventually reduces the number of features by a hybrid procedure and finally comes up with the optimal feature space. Hence, the model works on both dimensions effectively which results in a better accuracy and improves the team of classifiers performance. So, the proposed method is an effective dimensionality reduction method that is able to work well on the wide range of different datasets.

## 10.4 Performance Evaluation and Analysis

From the figure 5, we can see that the average number of misclassified samples for the proposed method is less than the other feature selection methods for all of the datasets. This shows that our proposed method is capable of improving the performance of the group of classification algorithms simultaneously and works well on different datasets with different sizes. As it is clear from the figure 6, the average relative absolute error of the group of classification algorithms for the proposed method is less than 10 for Wine, WDBC and Dermatology datasets. The figure also shows that the average classification error of the proposed method is less than 3 to 6 times comparing with the rest of the feature selection methods. This fact shows that our proposed method has decreased the classification error of the group of classifiers considerably. From the figure 7, the average TPRate of the group of classification algorithms for the proposed method is above 0.9. This shows that the proposed method increases the rate of true prediction considerably.

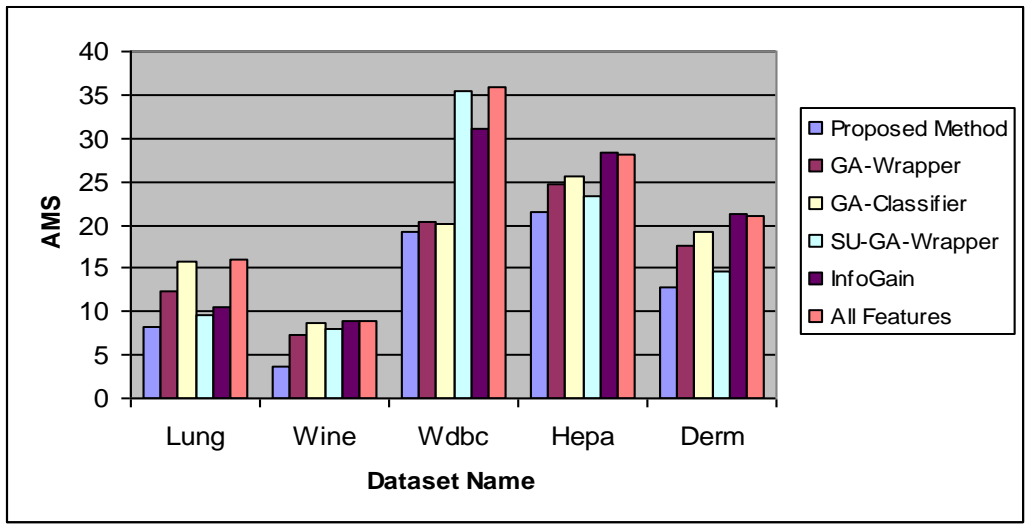

**Fig 5: Accuracy comparison between different methods by evaluation of AMS parameter**

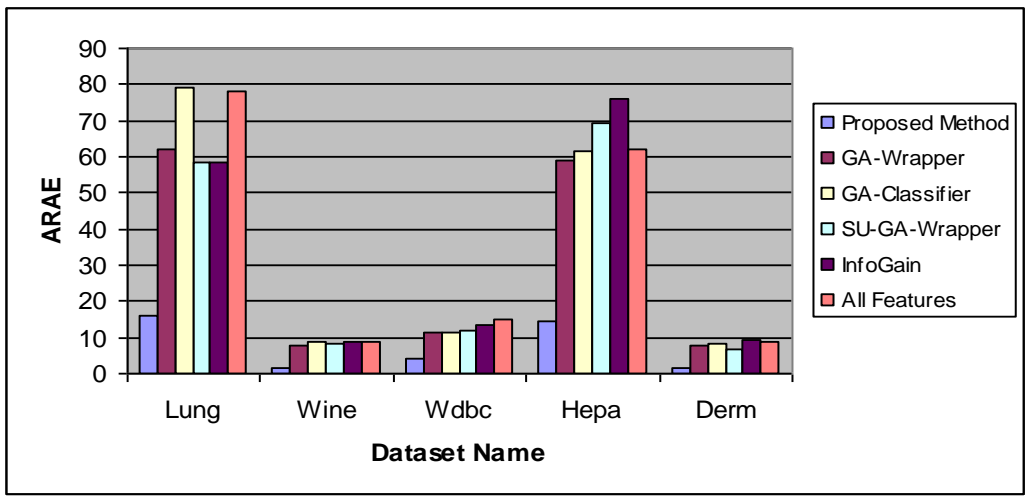

**Fig 6: Accuracy comparison between different methods by evaluation of ARAE parameter**

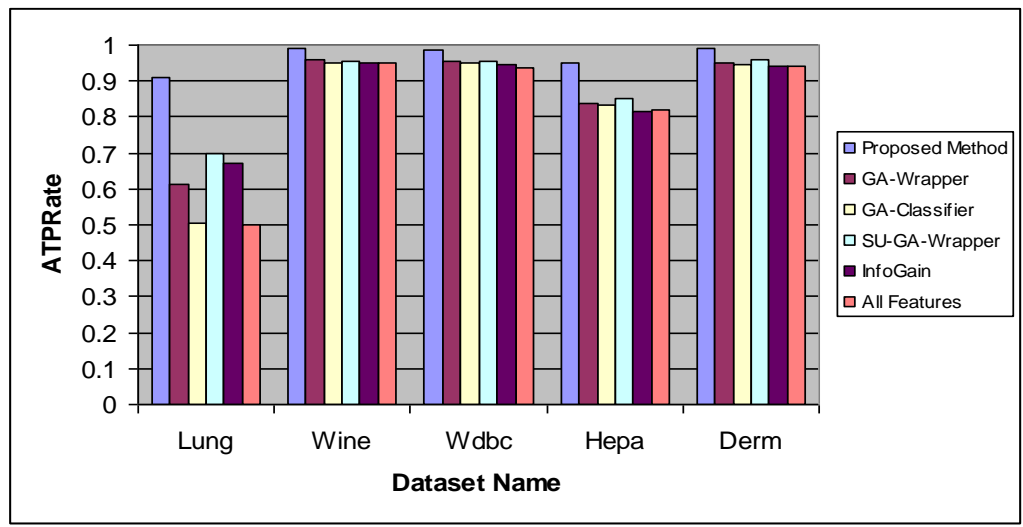

**Fig 7: Accuracy comparison between different methods by evaluation of ATPRate parameter**

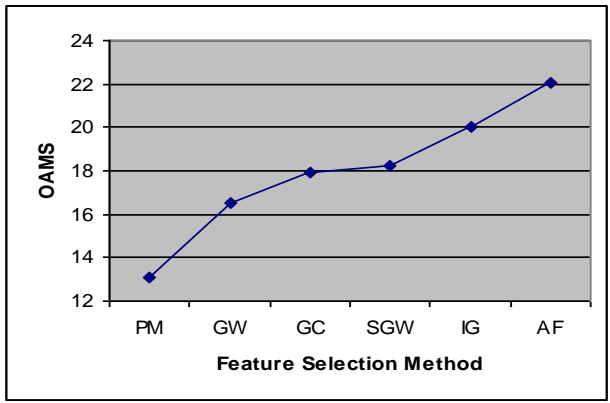

**Fig 8: OAMS parameter values for different methods**

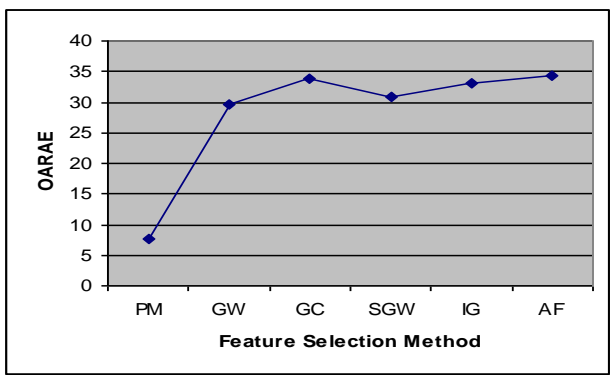

**Fig 9: OARAE parameter values for different methods**

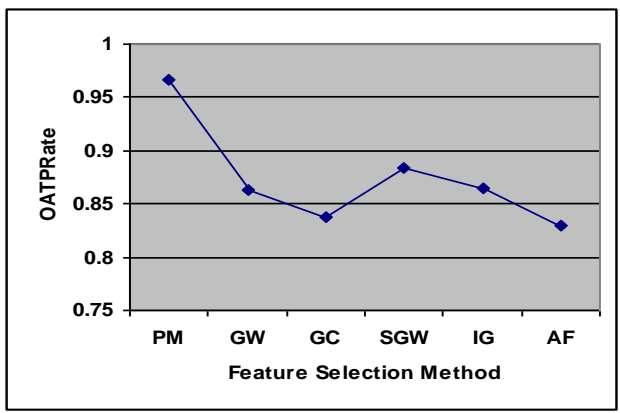

**Fig 10: OATPRate parameter values for different methods**

In the figure 8, we can see that the overall average number of misclassified samples in the proposed method for the group of classifiers in different datasets is less than other feature selection methods (GA-Wrapper, GA-Classifier, Symmetrical Uncertainty-GA-Wrapper, Information Gain, All Features). This shows that the proposed method is able to improve the accuracy of the group of classification algorithms on different datasets with various sizes. In figure 9, the overall average absolute error of the group of classifiers which are applied on 5 datasets (Lung cancer, WDBC, Hepatitis, Dermatology, Wine) are shown. It is clear from the figure that the proposed method's OARAE parameter is less than 4 times comparing with other feature selection methods. This means that the proposed feature selection method performance is much more higher than other methods. In the figure 10, the overall average true prediction rate of the group of classification algorithms which are applied on 5 datasets are shown. The true prediction rate is above 0.95. This shows that the proposed method has increased the true prediction rate of the group of classification algorithms on different datasets. The proposed method achieves higher classification accuracy for the group of classification algorithms in comparison to other methods. Moreover, the cost of our proposed method is considerably smaller than the GA-Wrapper and GA-Classifier methods. Furthermore, our proposed method works well on different datasets with various sizes.

## 11. CONCLUSION

In this paper, a hybrid two-phased feature selection method is proposed. This method takes advantages of mixing resampling and sample filtering with feature space filtering and wrapper methods. The first phase analyses sample domain and refines the samples to take the best result in the second phase. For the second phase, feature space filtering eliminates irrelevant features and then wrapper method select reliable features with a lower cost and higher accuracy. Different performance evaluation parameters are defined and calculated. The results show that our proposed method outperforms other feature selection methods (GA-Wrapper, GA-Classifier, Symmetrical Uncertainty-GA-Wrapper, Information Gain, All Features) on different datasets with different sizes. Furthermore, the proposed method improves the accuracy and true prediction rate of the group of classification algorithms simultaneously.

## 12. REFERENCES

[1] Silberschatz, A., Korth, H. F., and Sudarshan, S., 2010. Database System Concepts, McGrawHill.

[2] Bellman, R., 1996. Adaptive Control Processes: A Guided Tour, Princeton University Press, Princeton.

[3] Ladha, L., Deepa, T., 2011. Feature Selection Methods and Algorithms, International Journal on Computer Science and Engineering, Volume 3, page(s). 1787-1790.

[4] Liu, H., and Zhao, Z., 2012. Manipulating Data and Dimension Reduction Methods: Feature Selection, Journal of Computational Complexity, Page(s) 1790-1800.

[5] Janecek, G.K., Gansterer, N., Demel, A., and Ecker, F., 2008. On the relationship between feature selection and classification accuracy, Journal of Machine Learning and Research. JMLR: Workshop and Conference Proceedings 4, Pages 90–105.

[6] Assareh, A., Moradi, M., and Volkert, L., 2008. A hybrid random subspace classifier fusion approach for protein mass spectra classification, Springer, LNCS, Volume 4973, Page(s) 1–11, Heidelberg.

[7] Hayward, J., Alvarez, S., Ruiz, C., Sullivan, M., Tseng, J., and Whalen, G., 2008. Knowledge discovery in clinical performance of cancer patients, IEEE International Conference on Bioinformatics and Biomedicine, USA, Page(s) 51–58.

[8] Dhiraj, K., Santanu Rath, K., and Pandey, A., 2009. Gene Expression Analysis Using Clustering, 3rd international Conference on Bioinformatics and Biomedical Engineering.

[9] Jiang, B., Ding, X., Ma, L., He, Y., Wang, T., and Xie, W., 2008. A Hybrid Feature Selection Algorithm: Combination of Symmetrical Uncertainty and Genetic Algorithms, The Second International Symposium on Optimization and Systems Biology, Page(s) 152–157, Lijiang, China, October 31– November 3.

[10] Zhou, J., Peng, H., and Suen, C., 2008. Data-driven decomposition for multi-class classification, Journal of Pattern Recognition, Volume 41, Page(s) 67 – 76.

[11] Fayyad U., Piatetsky-Shapiro, G., and Smyth, P., 1996. From Data Mining A Knowledge Discovery in Databases, American Association for Artificial Intelligence.

[12] Novakovic, J., 2010. The Impact of Feature Selection on the Accuracy of Naïve Bayes Classifier, 18th Telecommunications Forum TELFOR, page(s) 1114-1116, November 23-25, Serbia, Belgrade.

[13] Domingos, P., Pazzani, M., 1997. On the Optimality of the Simple Bayesian Classifier under Zero-One loss, Machine Learning, Volume29, page(s) 103-130, November/December 1997.

[14] Blum, A.L., Rivest, R.L., 1992. Training a 3-node neural networks is NP-complete, Neural Networks, Volume 5, page(s) 117-127.

[15] Almuallim, H., Dietterich, T.G., 1991. Learning with many irrelevant features, in proceedings of AAAI-91, page(s) 547-552, Anaheim, California.

[16] Blum, A.L., Langley, P., 1997. Selection of Relevant Features and Examples in Machine Learning, Artificial Intelligence, Volume 97, page(s) 245-271.

[17] Chawla, N.V., Bowyer, K.W., Hall, L.O., and Kegelmeyer, W.P., 2002. SMOTE: Synthetic Minority Over-sampling Technique, Journal of Artificial Intelligence Research, Volume 16, page(s) 321-357.

[18] Kohavi, R., John, G.H., 1997. Wrappers for Feature Subset Selection, Artificial Intelligence, Volume 97, page(s) 273-324.

[19] Olusola, A.A., Oladele, A.S., Abosede, D.O., 2010. Analysis of KDD'99 Intrusion Detection Dataset for Selection of Relevance Features, Proceedings of the World Congress on Engineering and Computer Science, October20-22, San Francisco, USA.

[20] Randy, H., and Haupt, S., 1998. Practical Genetic Algorithms, John Wiley and Sons.

[21] Mertz C.J., and Murphy, P.M., 2013. UCI Repository of machine learning databases, http://www.ics.uci.edu/~mlearn/MLRepository.html, University of California.

[22] Dash, M., Liu, H., 2003. Consistency-based Search in Feature Selection, Artificial Intelligence, Volume 151, page(s) 155-176.

[23] Chou, T.S., Yen, K.K., and Luo, J., 2008. Network Intrusion Detection Design Using Feature Selection of Soft Computing Paradigms, International Journal of Information and Mathematical Sciences, Volume 4, page(s) 196-208.